\documentclass{article}
\pdfoutput=1

\usepackage[preprint, nonatbib]{neurips_2019}

\usepackage[utf8]{inputenc} % allow utf-8 input
\usepackage[T1]{fontenc}    % use 8-bit T1 fonts
\usepackage{hyperref}       % hyperlinks
\usepackage{url}            % simple URL typesetting
\usepackage{booktabs}       % professional-quality tables
\usepackage{amsfonts}       % blackboard math symbols
\usepackage{nicefrac}       % compact symbols for 1/2, etc.
\usepackage{microtype}      % microtypography

\usepackage{xcolor}
\usepackage{amsmath}
\usepackage{breqn}
\usepackage{float}
\usepackage{graphicx}
\usepackage{caption}
\usepackage{subcaption}
\usepackage[cal=cm]{mathalfa}

\usepackage[linesnumbered,ruled]{algorithm2e}
\usepackage[noend]{algpseudocode}
\usepackage[normalem]{ulem}
\makeatletter

\def\algbackskip{\hskip-\ALG@thistlm}
\makeatother

\usepackage{algorithm2e}

\usepackage{amsmath}
\usepackage{amssymb}
\usepackage{amsthm}

\usepackage{graphicx}
\usepackage{subcaption}
\usepackage{comment}
\usepackage{color}
\usepackage{etoolbox}
\newbool{inccomment}
\setbool{inccomment}{true}
\newcommand{\hl}[1]{\ifbool{inccomment}{{\color{magenta}#1}}{}}
\newcommand{\qxm}[1]{\ifbool{inccomment}{{\color{blue}#1}}{}}

\title{Temporal Surrogate Back-propagation for Spiking Neural Networks}

\author{
	Yukun Yang \\
	ECE Department\\
	UC Santa Barbara\\
	Santa Barbara, CA 93106 \\
	\texttt{yukunyang@ucsb.edu} \\
}

\begin{document}

\maketitle

\begin{abstract}
Spiking neural networks~(SNN) are usually more energy-efficient as compared to Artificial neural networks~(ANN), and the way they work has a great similarity with our brain. Back-propagation~(BP) has shown its strong power in training ANN in recent years. However, since spike behavior is non-differentiable, BP cannot be applied to SNN directly. Although prior works demonstrated several ways to approximate the BP-gradient in both spatial and temporal directions either through surrogate gradient or randomness, they omitted the temporal dependency introduced by the reset mechanism between each step. In this article, we target theoretical completion and investigate the effect of the missing term thoroughly. By adding the temporal dependency of the reset mechanism, the new algorithm is more robust to learning-rate adjustments on a toy dataset but does not show much improvement on larger learning tasks like CIFAR-10. Empirically speaking, the benefits of the missing term are not worth the additional computational overhead. In many cases, the missing term can be ignored.

\end{abstract}

\section{Introduction}
SNN is believed as the 3\textsuperscript{rd} generation artificial intelligence, its potential is yet to be discovered. Despite the high power efficiency as compared with traditional ANN, the application is constrained by the low accuracy brought by SNN's non-continuous spike behavior. How to boost SNN's performance and find its suitable application scenarios are the main focuses of current researches.

%while the baseline method has its after-defense ASR fluctuating between $2.4\%\sim51.4\%$.

%%% Local Variables:
%%% mode: latex
%%% TeX-master: "main"
%%% End:

\section{Background}
\label{sec:background}

\subsection{The Spiking Neuron Model}
The discrete time leaky integrate-and-fire~(LIF) model follows:
\begin{equation}
U^{(l+1)}[t+1] = U^{(l+1)}[t]\left(1-\frac{1}{\tau_m}\right)\left(1-S^{(l+1)}[t]\right)+\mathbf{W}^{(l)}a^{(l)}[t]
\label{eq:1}
\end{equation}

\begin{equation}
a^{(l+1)}[t+1] = a^{(l+1)}[t]\left(1-\frac{1}{\tau_s}\right)+\frac{1}{\tau_s}S^{(l+1)}[t+1]
\label{eq:2}
\end{equation}

\begin{equation}
    S^{(l+1)}[t] = H\left(U^{(l+1)}[t]-\vartheta\right), H(x)=\begin{cases}
        0,~ x<0 \\ 1,~ x>0
    \end{cases}
    \label{eq:3}
\end{equation}

\begin{figure}[t]
    \centering
    \includegraphics[width=\textwidth]{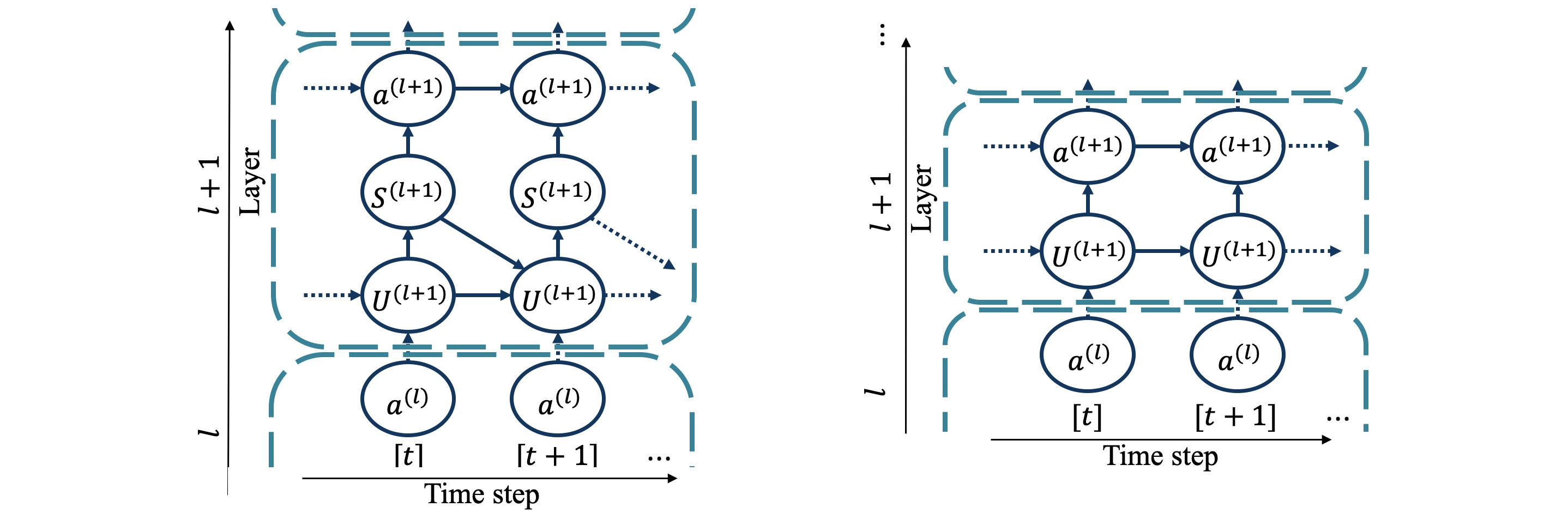}
    \caption{The two compute graphs}
    \label{fig:compute_graph}
\end{figure}

The compute graph is shown on the left side of Figure~\ref{fig:compute_graph}. We can safely substitute the variable $S$ by bringing the equation ~(\ref{eq:3}) into equations~(\ref{eq:1}) \& ~(\ref{eq:2}) and get:
\begin{equation}
    U^{(l+1)}[t+1] = U^{(l+1)}[t]\left(1-\frac{1}{\tau_m}\right)\left[1-H\left(U^{(l+1)}[t]-\vartheta\right)\right]+\mathbf{W}^{(l)}a^{(l-1)}[t]
    \label{eq:4}
\end{equation}

\begin{equation}
    a^{(l+1)}[t+1] = a^{(l+1)}[t]\left(1-\frac{1}{\tau_s}\right)+\frac{1}{\tau_s}H\left(U^{(l+1)}[t+1]-\vartheta\right)
    \label{eq:5}
\end{equation}

This leads our compute graph into the right side of Figure~\ref{fig:compute_graph}

\subsection{The Loss Function}
Use Van Rossum distance with kernel $\epsilon$ to measure the difference between a desired output spike trains $\boldsymbol{d}$ and an actual output spike trains $\boldsymbol{S}$:
\begin{equation}
    L = \sum_{k=0}^{N_t}E_{t_k} = \sum_{k=0}^{N_t}\frac{1}{2}\left((\epsilon\ast\boldsymbol{d})[t_k]-(\epsilon\ast\boldsymbol{S})[t_k]\right)^2
    \label{eq:6}
\end{equation}
%%% Local Variables:
%%% mode: latex
%%% TeX-master: "main"
%%% End:

\section{Methods}
\label{sec:methods}
According to the computation graph without variable $S$, and follows the basic Back-propagate through time (BPTT) strategy, we can get:
\begin{equation}
    \frac{\partial L}{\partial U^{(l)[t]}}=\underbrace{\frac{\partial L}{\partial a^{(l)[t]}}\frac{\partial a^{(l)[t]}}{\partial U^{(l)[t]}}}_{\textbf{Spatial}}+\underbrace{\frac{\partial L}{\partial U^{(l)[t+1]}}\frac{\partial U^{(l)[t+1]}}{\partial U^{(l)[t]}}}_{\textbf{Temporal}}
\label{eq:7}
\end{equation}
The activation based methods like SLAYER\cite{shrestha2018slayer}, STBP\cite{wu2018spatio}, SuperSpike\cite{zenke2018superspike}, and ANTLR \cite{kim2020unifying} choose different approaches to approximate the \textbf{Spatial} term. However, as for the \textbf{Temporal} part, we derived it based on equation~(\ref{eq:4}) as following:
\begin{equation}
    \frac{\partial U^{(l)[t+1]}}{\partial U^{(l)[t]}} = \left(1-\frac{1}{\tau_m}\right)\left\{\left[1-H\left(U^{(l)}[t]-\vartheta\right)\right] +  \color{red}{U^{(l)}[t]\left[-\frac{\partial H\left(U^{(l)}[t]-\vartheta\right)}{\partial U^{(l)}[t] }\right]}\color{black}{}\right\}
\end{equation}
The red colored term is missed in all previous works. This term is also ill-issdefined as the term ${\partial a^{(l)[t]}}/{\partial U^{(l)[t]}}$ in the \textbf{Spatial} term for the same reason: The derivative of $H(x)$ over $x$ is the Dirac delta function, which is zero almost everywhere and is infinity only when $x=0$. However this term encodes an important dependency of membrane potential between different time steps.
Like all previous works, we use the surrogate gradient method to approximate the Heaviside step function's gradient:
\begin{equation}
    H\left(U^{(l)}[t]-\vartheta\right) \approx \sigma\left.\left(U^{(l)}[t]-\vartheta\right)\right|_T = \frac{1}{1+e^{\frac{-\left(U^{(l)}[t]-\vartheta\right)}{T}}}
    \label{eq:9}
\end{equation}
    
\begin{equation}
    \frac{\partial H\left(U^{(l)}[t]-\vartheta\right)}{\partial U^{(l)}[t] } \approx \frac{1}{T} \left.\sigma\left(U^{(l)}[t]-\vartheta\right)\right|_T * \left(1-\sigma\left.\left(U^{(l)}[t]-\vartheta\right)\right|_T\right)
    \label{eq:10}
\end{equation}
The new surrogate gradient term describes the timing dependency brought by the reset mechanism: When the membrane potential near the threshold, small perturbation towards threshold has potential to change the membrane potential of the next time-step dramatically.
%%% Local Variables:
%%% mode: latex
%%% TeX-master: "main"
%%% End:
\section{Experiments}
\label{sec:experiments}

\begin{figure}[t]
    \centering
    \includegraphics[width=\textwidth]{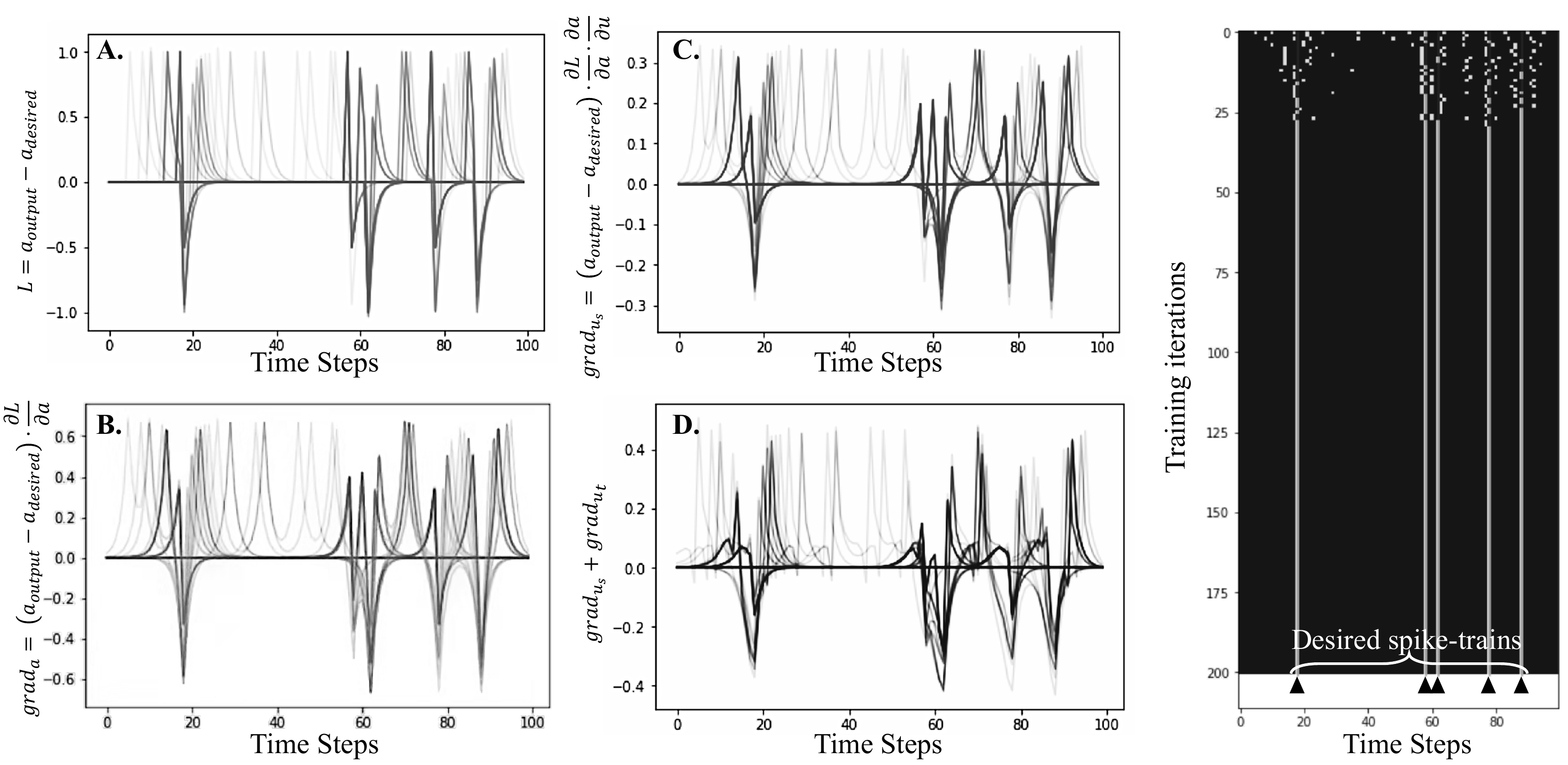}
    \caption{Single neuron example, and corresponding different compute phases of BP gradient}
    \label{fig:toy_experiment}
\end{figure}

\subsection{Toy Experimental Setup}
\label{sec:toy_setup}
We simulate a single neuron, with fixed random input spike current and fixed fictitious random target spike-trains. We use standard BP algorithm to train the neuron by 200 iteration. All parameters are concluded in the Table~\ref{table:1}:\par
\begin{table}[h]
    \centering
    \caption{Important parameters of the single neuron back-propagation experiment}
    \begin{tabular}{|c|c|c|c|c|c|c|c|} 
        \hline 
        \#Input neuron&50&Desired spike-train $p$&1/20&$\tau_m$&6&Sigmoid $T$&0.3\\
        \hline  
        \#Time steps&100&Input spike-train $p$&1/10&$\tau_s$&2&Default lr&0.005\\
        \hline 
    \end{tabular}
    
    \label{table:1}
    
\end{table}

In the Table~\ref{table:1}, variable $p$ means the probability of generating a spike on each time-step when initialize fictitious inputs or a target output spike-trains.
As shown in the right side of Figure~\ref{fig:toy_experiment}, the neuron's output converged to the desired output successfully near the 30th iteration.\par
On the left side, we put four figures to show the four computing phases of gradients (All 200 iterations overlapped together). The Figure A is the difference between the current and the desired output. Then the difference is propagated to spike-trains in the Figure B. Since all previous spikes contribute to later current, the new graph gains a left tail on each spike compared to Figure A. By multiply the surrogate gradient of $\partial S/\partial U$, the Figure B transit to the $grad_{u_s}$ shown in Figure C. Finally, the Figure D add the timing dependency with the missing term $grad_{u_t}$ mentioned above onto the $grad_{u_s}$. \par

\subsection{Toy Experiment Results}
\begin{figure}[t]
    \centering
    \includegraphics[width=\textwidth]{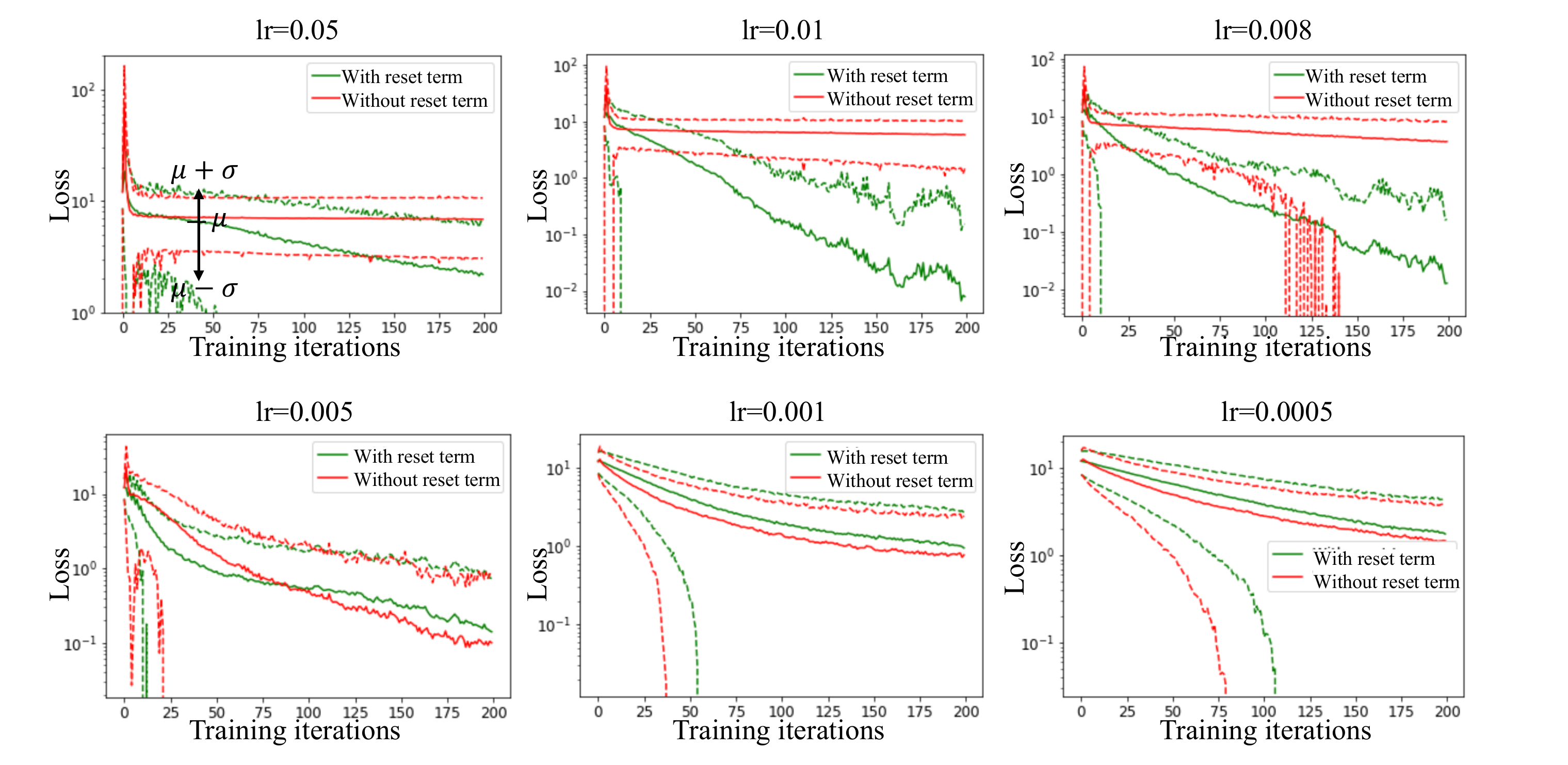}
    \caption{Toy example comparison between with/without the reset dependency}
    \label{fig:toy_compare}
\end{figure}

We repeat this toy experiment multiple times and record the loss curve. Figure~\ref{fig:toy_compare} compares the loss curve before and after adding the reset term. The solid lines are averaged losses, and the dash lines are one standard deviations above/below the averages. After adding the reset term, the algorithm performs better when learning rate is $>0.005$ as shown on the first line of Figure~\ref{fig:toy_compare}, and has competitive performance with smaller learning rate.

\subsection{Larger datasets}
We conclude our current results in the Table~\ref{table:2}. The parameters of the CNNs we used are shown in the Table~\ref{table:3} below: 
\vspace{-0.3cm}
\begin{table}[h]
    % \vspace{-1cm}
    \centering
    \caption{Performance comparison on larger datasets}
    \begin{tabular}{c|c|c} 
        % \hline 
        Datasets& With reset term accuracy(\%) & Without reset term accuracy(\%)\\
        \hline  
        NMNIST&99.29&99.28\\
        MNIST&99.49&99.50\\
        CIFAR10& 88.96 & 88.98\\
        % \hline 
    \end{tabular}
    \label{table:2}
\end{table}
\vspace{-0.5cm}
\begin{table}[h]
    % \vspace{-1cm}
    \centering
    \caption{CNN's parameters}
    \begin{tabular}{c|c|c} 
        % \hline 
        Datasets& Network Size & Time steps\\
        \hline  
        NMNIST & 12C5-P2-64C5-P2 & 5\\
        MNIST & 15C5-P2-40C5-P2-300 & 30\\
        CIFAR10 & 96C3-256C3-P2-384C3-P2-384C3-256C3-1024-1024 & 5\\
        % \hline 
    \end{tabular}
    \label{table:3}
\end{table}

%%% Local Variables:
%%% used is mode: latex
%%% TeX-master: "main"
%%% End:
\section{Conclusion and future works}
\label{sec:conslusion}

In this short article. We add on the previous missed reset dependency between time steps using surrogate gradient. The new term shows better performance on the single neuron experiment when learning rate is larger than 0.005. However, on real datasets include MNIST, NMNIST and CIFAR-10, adding this reset dependency does not help.\par
% In the future, we will investigate the effect of the decay factors:
% \begin{equation}
%     D_s = (1-\frac{1}{\tau_s})
% \end{equation}
% \begin{equation}
%     D_m = (1-\frac{1}{\tau_m})
% \end{equation}
% Instead of using a unified decay factor on the whole network, we will set each neuron with a decay factor different from each others, and make this variable adjustable during BP. More aggressively speaking, the decay factors can even larger than one, then a neuron will amplify an input signal. Maybe this additional timing non-linearity will bring some benefits to SNN's performance.
%%% Local Variables:
%%% mode: latex
%%% TeX-master: "main"
%%% End:

\bibliographystyle{unsrt}
\bibliography{reference}

\begin{thebibliography}{1}

\bibitem{shrestha2018slayer}
Sumit~B Shrestha and Garrick Orchard.
\newblock Slayer: Spike layer error reassignment in time.
\newblock {\em Advances in Neural Information Processing Systems}, 2018.

\bibitem{wu2018spatio}
Yujie Wu, Lei Deng, Guoqi Li, Jun Zhu, and Luping Shi.
\newblock Spatio-temporal backpropagation for training high-performance spiking
  neural networks.
\newblock {\em Frontiers in neuroscience}, 12:331, 2018.

\bibitem{zenke2018superspike}
Friedemann Zenke and Surya Ganguli.
\newblock Superspike: Supervised learning in multilayer spiking neural
  networks.
\newblock {\em Neural computation}, 30(6):1514--1541, 2018.

\bibitem{kim2020unifying}
Jinseok Kim, Kyungsu Kim, and Jae-Joon Kim.
\newblock Unifying activation-and timing-based learning rules for spiking
  neural networks.
\newblock {\em Advances in Neural Information Processing Systems}, 2020.

\end{thebibliography}

\newpage
\normalsize
\appendix

\end{document}